# Traffic4cast 2020 – Graph Ensemble Net and the Importance of Feature & Loss Function Design for Traffic Prediction

**Qi Qi**
qiq208@gmail.com

**Pak Hay Kwok**
pak_hay_kwok@hotmail.com

## Abstract

This paper details our experience in, and our solution to Traffic4cast 2020. Similar to Traffic4cast 2019, Traffic4cast 2020 challenged its contestants to develop algorithms that can predict the future traffic states of big cities. Our team tackled this challenge on two fronts. We studied the importance of feature and loss function design, and achieved significant improvement to the best performing U-Net solution from last year. We also explored the use of Graph Neural Networks (GNNs) and introduced a novel ensemble GNN architecture which outperformed last year's GNN implementations. While our GNN was improved, it was still unable to match the performance of U-Nets and the potential reasons for this shortfall were discussed. Our final solution, an ensemble of our U-Net and GNN, achieved the 4th place solution in Traffic4cast 2020.

## 1 Introduction

Traffic4cast 2020 was a competition which challenged its contestants to develop algorithms that can predict the future traffic states of big cities. Similar to Traffic4cast 2019 [1], this year's competition was built upon real traffic data from Berlin, Moscow and Istanbul. The data was in the form of a sequence of images containing traffic aggregated in 100m x 100m x 5min spatio-temporal time bins. Different to last year, any proposed algorithm would need to provide prediction up to an hour into the future, instead of 15 minutes.

Various flavours of U-Net dominated last year's competition, with both the top 2 teams, Choi [2] and Martin et al. [3], adopting U-Net in their final solutions. Furthermore, Martin et al. [3] evaluated a range of architectures and selected their U-Net because of its superior performance. Hence, after reviewing last year's solutions and some preliminary experimentations, we decided to focus on two areas of investigation:

- Importance of feature and loss function design, and other non-architectural changes

  The dominance of U-Nets in last year's competition demonstrated their suitability for the task. Therefore, we decided to build upon last year's U-Net solutions and seek improvements by conducting experiments on the choice of input features, loss functions, learning rate schedules, etc.

- Effectiveness of GNNs

  GNNs facilitate the application of deep learning techniques and ideas to data with an underlying structure that is non-Euclidean. While the competition traffic data was processed into a regular image structure, we believed that road networks were natively more aligned to graph representations. This is because the ability of the traffic on one road to affect (or pass information onto) another is dependent on whether they are closely connected and not on their spatial proximity. For example, a small country road may weave close to a controlled-access highway. If the highway has no exchanges in the region, then these two roads are highly unlikely to share traffic characteristics even though they are spatially in proximity. The intuition is that a GNN in which the edges capture the road connectivity will inherently adhere to these physical constraints and also allow information to propagate over greater spatial distance. Moreover, there have been significant developments in GNNs recently [4, 5], which led us to explore GNNs and their advancements.

Our modifications to the input features, loss functions and learning rate schedules yielded a U-Net solution which was better than last year's top solution. While the investigation into GNNs resulted in a novel ensemble type architecture, which leverages three popular, yet differing, graph convolution kernels that have recently been developed (ChebConv by Defferrard et al. [6], SAGEConv by Hamilton et al. [7] and SGConv by Wu et al. [8]). Although the proposed GNN could not compete with the improved U-Net, it did outperform last year's GNN solution by Martin et al. [9]. Our final solution was an ensemble of our U-Net and GNN and it delivered the 4th place solution.

## 2 Methods

### 2.1 U-Net

The architecture of our baseline U-Net is shown in Figure 1, it largely resembles last year's top performing U-Net by Choi [2] and it served as a platform for our experiments on features and loss functions. Modifications were introduced mainly around the input and output layers in order to handle the increased number of input and output channels. This architecture was chosen after some preliminary experiments on alternatives, such as the same U-Net but with fewer hidden layers, a 3D U-Net, and a nested U-Net (U-Net++) by Zhou et al. [10]. While U-Net++ did provide incremental improvement, it was not selected due to a substantial increase in the number of parameters and thus the memory usage. The chosen U-Net remained as the most balanced option considering both performance and memory requirement.

The final U-Net for submission was trained using Cyclic Cosine Annealing described by Loshchilov et al. [11]. As the name suggested, training was split into cycles with each cycle consisting of 7 epochs. At each cycle, the learning rate was first reset to a maximum learning rate of 3e-4, then it was reduced following a cosine annealing schedule. Resetting the learning rate back to the maximum at the beginning of each cycle perturbs the model and encourages it to explore multiple basins of attractions. Training was continued until an additional cycle failed to return a better validation score. Once the training was stopped, the final model weights were calculated using Snapshot Ensemble proposed by Huang et al. [12]. From the training history, the best snapshot model from each cycle was compared, and the model weights of the top 3 snapshot models were averaged to give the final model.

### 2.2 Graph Neural Network

#### 2.2.1 Graph generation

To apply graph-based techniques, a road network needed to be reconstructed from the image data. As such, a single image was created, which stored the maximum historical traffic volume across all directions per pixel. Theoretically, any pixel which had a non-zero maximum would represent a pixel containing a road. However, this resulted in a very densely connected graph, and would therefore reduce the effectiveness of GNN approaches. To alleviate this issue, we added a threshold value of 5, whereby any pixel with a maximum historical volume of less than 5 was ignored in the road network (Fig. 2).

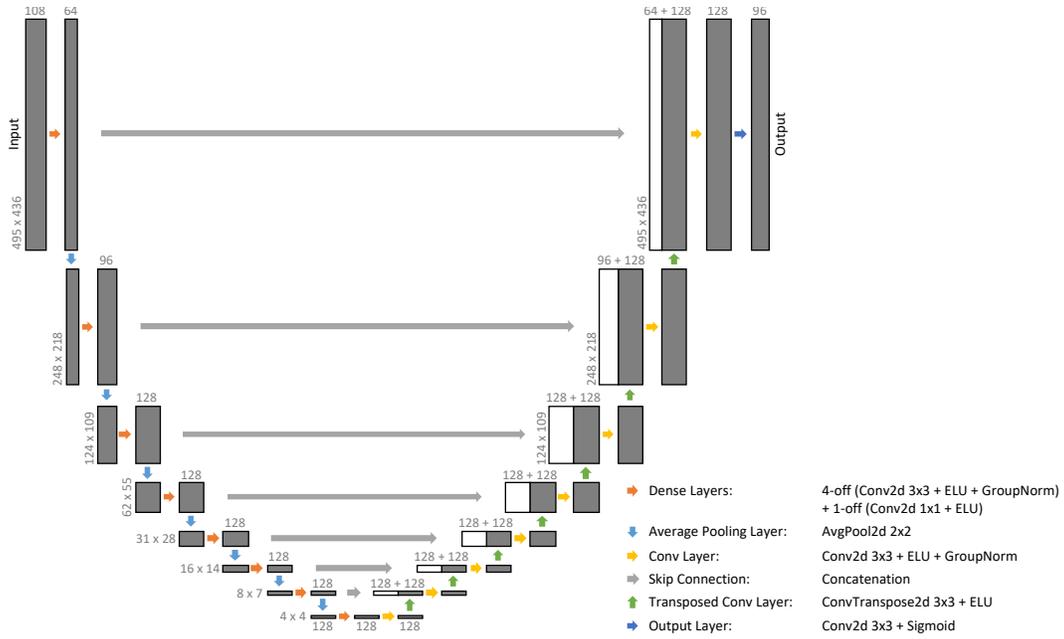

Figure 1: The U-Net architecture.

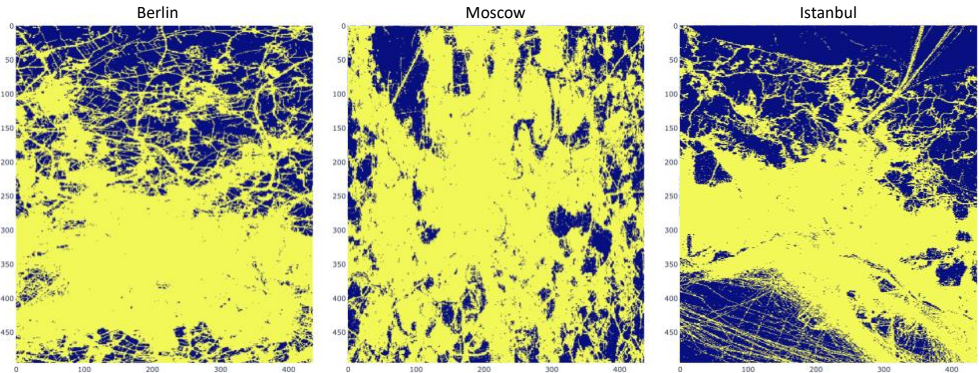

Figure 2: The assumed 'road network' (pixels with max volume above 5).

### 2.2.2 Graph Ensemble Net

The explosion of interest in GNNs has resulted in the development of a vast number of approaches to implement convolutions on graphs [4, 5]. Currently, the PyTorch Geometric library by Fey et al. [13], which was used for this competition, contains nearly 40 different types of convolutional layers, with many more in the pipeline. The designs of these filters borrow their ideas from a wide range of different fields (e.g. RNN, CNN, spectral graph theory, etc.) and so approach the problem from slightly different angles. Therefore, we decided to see if we could leverage this diversity by applying several differing convolutions and ensembling them within the training itself.

To implement this, we adapted the Graph ResNet from last year's competition and replaced the 'Graph res block' with our new ensemble block. In this block, the inputs are fed into each convolutional filter, with the outputs averaged before being passed on to the next layer/block.

We needed the convolutional filters within the block to be as different as possible, and also as simple as possible to ensure acceptable memory and runtime requirements. Thus, we selected ChebConv [6] to capture spectral-based insights, SAGEConv [7] to capture local neighbours' insights and SGConv [8] to capture broader neighbourhood insights (with K set to 5).

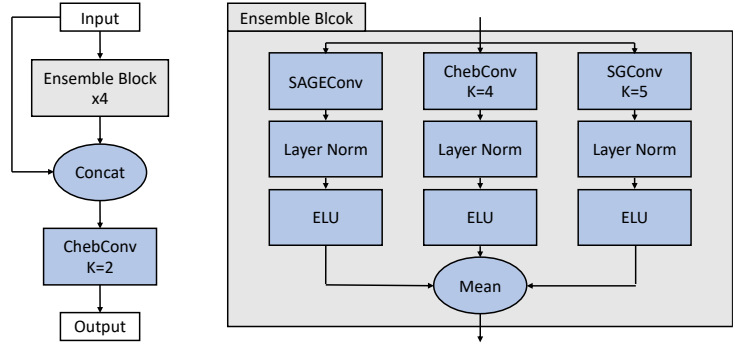

Figure 3: The Graph Ensemble Net architecture.

## 2.3 Inputs

While the input traffic data was three-dimensional in nature, both our U-Net and GNN were two-dimensional methods. To resolve this dimensional incompatibility, we collapsed the temporal dimension of the inputs into channels, resulting in inputs of shape (495, 436, 108) for U-Net and shape (Number of nodes, 108) for GNN. Other input transformations, summarised in Table 1, were applied, resulting in final input shapes of (495, 436, 143) for U-Net and shape (Number of nodes, 91) for GNN.

Table 1: Summary of input transformations

| Transformation | Description | Applied to | Change to No. of Channels |
| --- | --- | --- | --- |
| Append city static features | Features containing the locations of junctions and landmarks, such as parking and transport | Both | +7 |
| Append time of day feature | Feature containing the starting time of day of the input | Both | +1 |
| Append aggregate features | Features containing the range, mean and standard deviation across the 12 input frames | U-Net | +(3*9) |
| Apply input mask | Mask pixels with maximum traffic volume less than 5 | U-Net | 0 |
| Normalisation | Divide input by 255 | U-Net | 0 |
| Append pixel coordinate feature | Features encoding pixel coordinates of nodes | GNN | +2 |
| Replace the first 6 frames by aggregate | Replace the first 6 frames by mean, min and max | GNN | -(3*9) |
| Standardisation | Standardise input by mean and standard deviation | GNN | 0 |

## 2.4 Outputs and loss function

The competition requested outputs of shape (6, 495, 436, 8) where the first dimension denoted the 6 forward prediction frames at 5, 10, 15, 30 and 60 minutes. While the trivial solution would be to configure the model to output just the 6 required frames, we felt that having a consistent, 5-minute spacing between output frames might better guide the model during training. Hence, our model outputs were of shape (12, 495, 436, 8) and the training loss function was the pixel-wise mean squared error (MSE) considering all 12 frames. The 6 extra frames were discarded during validation and testing.

Another intuition was that the maximum volume and speed of traffic per direction per pixel across the available data represented a hard constraint which was unlikely to be exceeded in the prediction frames. Thus, a single number was calculated per pixel per channel containing the maximum historical value. This was then used to clamp the final predictions.

# 3 Results

Aside from the leaderboard scores listed in Table 2, all other experiments were conducted using the Berlin dataset only, with scores quoted as the validation pixel-wise MSE. We found excellent correlation between the Berlin validation score and the leaderboard score.

## 3.1 Leaderboard

Table 2: Final leaderboard scores

| Model | Overall Leaderboard Score |
|---|---|
| Mean Baseline | 0.0013265 |
| Graph Ensemble Net | 0.0012107 |
| U-Net | 0.0011768 |
| U-Net + Graph Ensemble Net | 0.0011764 |

Table 2 listed the leaderboard scores of our models relative to the mean baseline. Whilst our U-Net outperformed our Graph Ensemble Net, further performance gain was realised by aggregating the predictions from both our models.

## 3.2 U-Net

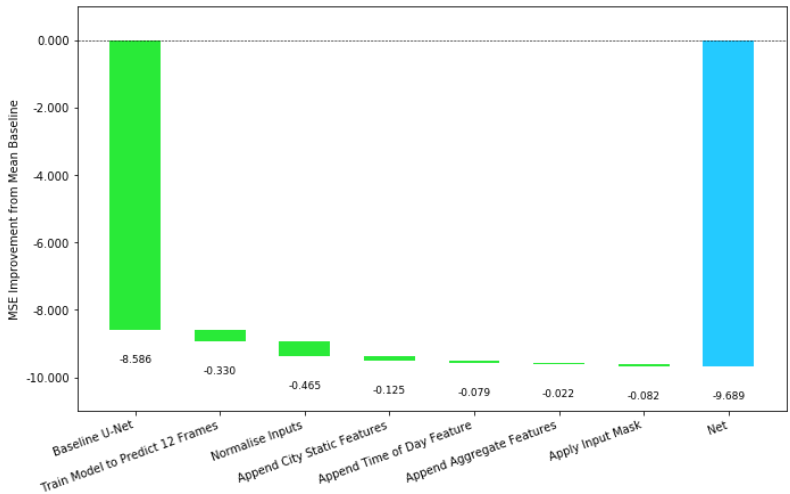

Figure 4: U-Net improvement breakdown.

Figure 4 illustrates the U-Net's 'journey of improvement', all values in the figure are relative to the Berlin mean baseline score of 93.720. As shown in the figure, while almost 90% of the total improvement were attributed to the introduction of the baseline U-Net, further incremental improvements were realised and these represented advancements from last year's top solution. Simple modifications, such as training the model to consider the losses of all 12 forward prediction frames instead of just the 6 required frames, and normalising the input yielded considerable improvements. Other modifications, such as the applications of the output clamp, snapshot ensemble, and model ensemble gave the remaining improvements from the Berlin experiments to the final submission.

## 3.3 Graph Ensemble Net

Table 3: Comparison of GNNs

| Model | Berlin Validation MSE |
|---|---|
| Mean Baseline | 93.720 |
| Graph ResNet | 88.930 |
| Graph Ensemble Net | 87.130 |

For the comparison of Graph ResNet, the same upstream and downstream feature engineering techniques were applied. Different learning rates schedules and batch sizes were tested to find the best Graph ResNet score without any architectural changes to the network.

## 4 Discussion

### 4.1 U-Net loss function

Out of all the improvements, the one attained by modifying the loss function to consider all 12 forward prediction frames was one of the most significant and by far the most unexpected. Given this finding, we believed that additional improvement would be possible by further refining the loss function. Indeed, one may consider replacing the loss function and use a generative adversarial network (GAN) instead. However, training a GAN requires iterative training of both a generator and a discriminator, substantially increasing the training time. Alternatively, one may calculate the 'feature reconstruction loss' by passing in both the prediction and the target to some pre-trained image network, such as VGG-16, as proposed by Johnson et al. [14]. Though the traffic maps were in the form of images, they were not conventional images, and the difference in context might therefore reduce the effectiveness of Johnson's method.

Instead of passing in both the prediction and target to some pre-trained image network, we proposed passing them through the encoder part of the U-Net and calculate the 'hidden layer losses'. Figure 5 is a graphical representation of the method. At the end of each layer, just prior to the average pooling operation, the hidden layer loss is calculated as the MSE between the current prediction and target activations. The final loss function of which the gradient is backpropagated is the weighted sum of the original pixel-wise MSE ($L_{h0}$) and the hidden layer losses ($L_{h1}$, $L_{h2}$, …, $L_{h8}$). The advantages of this method are that it does not rely on any external discriminator or network and that the hidden layer losses will always fit the context of the problem which is being solved.

Initial testing of this proposed method showed promising results, especially on its ability in suppressing the hidden layer losses from growing. However, it was not included in the final solution given the limited time of the competition.

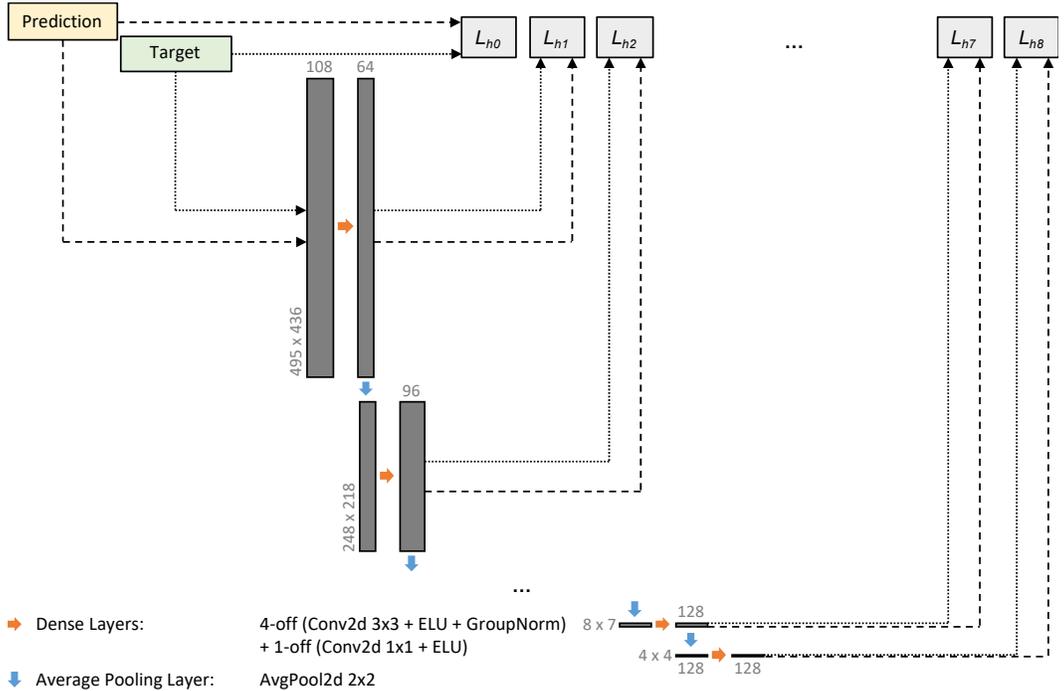

Figure 5: Hidden layer losses.

## 4.2 Graph Ensemble Net

We explored several different architectures, convolutional filters, and pooling methods in our exploration of a GNN solution. We found that the more complex filters did not offer any performance benefits and only significantly slowed down training and increased memory usage.

## 4.3 GNN vs U-Net

Whilst the Graph Ensemble Net architecture provided better than the mean baseline and improved on the GNN architecture from last year, it still could not compete with the U-Net solution.

It is hypothesised that this is due to the fact the data is split into an aggregation of traffic over 100m x100m cells which essentially smoothed out road-level data (Fig. 2). Given that the data source is from 'a large fleet of probe vehicles', i.e. there seem to be measurements taken on actual roads, we wonder if the raw data could be mapped and aggregated onto a real road network. It would be interesting to see under these conditions, whether graph-based methods could become competitive.

We wondered how these predictions would be used to provide actionable insights towards the overall mission of improving mobility and sustainability. If the predictions are to be used to inform road-level interventions then perhaps more granular, road-level predictions would be beneficial (and even necessary) over the current 100m x100m regional insights.

If this is the case, the road-level granularity will generate too large of a dataset to handle, then perhaps a hybrid multi-fidelity U-Net-GNN approach could be tried. Here a U-Net could make the macro prediction highlighting regions of interest, where a GNN can then be deployed to produce road-level predictions.

## 5 Conclusions

Traffic4cast 2020 allowed us to further improve the prediction capability of some of the methods developed in last year's competition. We approached the problem from two different angles which led to some interesting insights, as well as prediction improvement. We saw again the suitability of U-Nets for this problem, and demonstrated significant improvement through feature and loss function design. We proposed a new loss function which might further improve the U-Net. We presented a novel ensemble GNN architecture which outperformed last year's GNN implementations but was still unable to compete with the U-Net solution. We hypothesised that the current 100m x 100m aggregation of data meant the problem was more suited for U-Nets and suggested alternative data aggregation might unlock the potential of GNNs.